\documentclass[aip,graphicx]{revtex4-1}

\usepackage[utf8]{inputenc}
\usepackage{xcolor}
\usepackage{hyperref}
\usepackage{amssymb}
\usepackage{amsmath}
\usepackage{lineno}
\usepackage{array}
\usepackage{caption}
\usepackage{subcaption}
\usepackage{multirow}
\usepackage{csquotes}
\usepackage{amsfonts}
\usepackage{relsize}
\usepackage{bbm}
\usepackage{comment}
\usepackage{graphicx}
\usepackage{makecell}
\usepackage{amsmath}

\draft 

\begin{document}

\preprint{Version 1}

\title{DropTrack - automatic droplet tracking using deep learning for microfluidic applications } 



\author{Mihir Durve}

\email{mihir.durve@iit.it}
\affiliation{Center for Life Nano- \& Neuro-Science, Fondazione Istituto Italiano di Tecnologia (IIT), viale Regina Elena 295, 00161 Rome, Italy}

\author{Adriano Tiribocchi}
\affiliation{Istituto per le Applicazioni del Calcolo del Consiglio Nazionale delle Ricerche, via dei Taurini 19, 00185, Rome, Italy}

\author{Fabio Bonaccorso}
\affiliation{Istituto per le Applicazioni del Calcolo del Consiglio Nazionale delle Ricerche, via dei Taurini 19, 00185, Rome, Italy}
\affiliation{Department of Physics and National Institute for Nuclear Physics, University of Rome "Tor Vergata'', Via Cracovia, 50, 00133 Rome, Italy}

\author{Andrea Montessori}
\affiliation{Dipartimento di Ingegneria, Università degli Studi Roma tre, via Vito Volterra 62, Rome, 00146, Italy}

\author{Marco Lauricella}
\affiliation{Istituto per le Applicazioni del Calcolo del Consiglio Nazionale delle Ricerche, via dei Taurini 19, 00185, Rome, Italy}

\author{Micha{\l} Bogdan}
\affiliation{Institute of Physical Chemistry, Polish Academy of Sciences, Kasprzaka 44/52,
01-224 Warsaw, Poland}

\author{Jan Guzowski}
\affiliation{Institute of Physical Chemistry, Polish Academy of Sciences, Kasprzaka 44/52,
01-224 Warsaw, Poland}

\author{Sauro Succi}
\affiliation{Center for Life Nano- \& Neuro-Science, Fondazione Istituto Italiano di Tecnologia (IIT), viale Regina Elena 295, 00161 Rome, Italy}
\affiliation{Istituto per le Applicazioni del Calcolo del Consiglio Nazionale delle Ricerche, via dei Taurini 19, 00185, Rome, Italy}
\affiliation{Department of Physics, Harvard University, 17 Oxford St, Cambridge, MA 02138, United States}


\date{\today}

\begin{abstract}
Deep neural networks are rapidly emerging as data analysis tools, often outperforming the conventional techniques used in complex microfluidic systems. One fundamental analysis frequently desired in microfluidic experiments is counting and tracking the droplets. Specifically, droplet tracking in dense emulsions is challenging as droplets move in tightly packed configurations. Sometimes the individual droplets in these dense clusters are hard to resolve, even for a human observer. Here, two deep learning-based cutting-edge algorithms for object detection (YOLO) and object tracking (DeepSORT) are combined into a single image analysis tool, DropTrack, to track droplets in microfluidic experiments. DropTrack analyzes input videos, extracts droplets' trajectories, and infers other observables of interest, such as droplet numbers. Training an object detector network for droplet recognition with manually annotated images is a labor-intensive task and a persistent bottleneck. This work partly resolves this problem by training object detector networks (YOLOv5) with hybrid datasets containing real and synthetic images. We present an analysis of a double emulsion experiment as a case study to measure DropTrack's performance. For our test case, the YOLO networks trained with 60\% synthetic images show similar performance in droplet counting as with the one trained using 100\% real images, meanwhile saving the image annotation work by 60\%. DropTrack's performance is measured in terms of mean average precision (mAP), mean square error in counting the droplets, and inference speed. The fastest configuration of DropTrack runs inference at about $30$ frames per second, well within the standards for real-time image analysis. 
\end{abstract}


\maketitle 

\section{Introduction}
\label{introduction}
Over the last decade, machine learning (ML) has become a popular and useful tool in most scientific and technological fields. Despite some limitations~\cite{SucciCoveney, riva2022}, deep learning-based ML tools can analyze massive datasets, make predictions, learn hidden patterns, draw inferences, or solve problems without being explicitly programmed for it~\cite{LeCun2015}. Recent advancements in ML techniques leveraging ability to analyze Big Data have led to many commercial applications such as handwriting recognition~\cite{darmatasia,ahlawat}, speech recognition~\cite{tandel,han2014speech}, text sentiment analysis of posts on social media platforms~\cite{severyn2015,ramadhani2017,zhang2018}, traffic management, customer service, and crowd management~\cite{osman2017,ho2019,yogameena,ragesh2019}, to name but a few. The state-of-the-art tools often surpass human-level performance in performing some of these tasks~\cite{niklas}. 

In the scientific domain, machine learning tools have proved to be indispensable assets in achieving recent milestones~\cite{Alfa_Fold, CARLEO}. Laboratories or large collaborative experiments are increasingly using machine learning tools to gain insights from the generated big data~\cite{yu2022,Coelho2021}. Currently, ML approaches are being adapted for domain-specific applications across various scientific disciplines aiming at discovering the fundamental principles~\cite{Thiyagalingam2022}. In the microfluidics domain, for example, Hadikhani et al.\cite{hadikhani} trained a neural network to predict fluid and flow properties using digital images of the experimental setup and Mahdi et al.\cite{mahdi} used neural networks to predict size distribution of the droplets in an emulsion. A convolutional autoencoder model was trained to discover a low-dimensional representation to describe droplet shapes within a concentrated emulsion and classify stable vs unstable droplets from their shapes~\cite{khor}.

Microfluidic applications often require fast and real-time monitoring, such as droplet counting, droplet tracking, the size distribution, and other properties of the droplets. In particular, automated tracking of moving objects is highly relevant in high-throughput microfluidic systems, which involve thousands of flowing droplets generated at very high frequencies~\cite{zhu2017}. Inferring these quantities using video analysis tools would reduce the need of specialized measuring hardware equipments~\cite{zamboni2021, schianti2019}. A digital camera connected to a microscope can conveniently be installed to observe small droplets within microfluidic chips and it would be totally sufficient for this purpose. However, the multi-stage image transformations in this setup add various types of noise at different 
stages~\cite{chen2009}. Moreover, in some microfluidically generated materials, such as dense emulsions~\cite{montessori2021_1, montessori2021_2}, the droplets undergo significant deformations, accelerated motions, and form densely packed configurations that are often difficult to resolve and track in a video, thus making classical image analysis tools insufficient to infer quantities of 
interest. Despite these challenges, the desired observables can be measured by the latest computer vision tools that are being used to track objects such as cars, etc., from traffic surveillance cameras~\cite{tao2017,lin2018}. These computer vision tool essentially perform two tasks, viz. identifying objects in an image and tracking the objects across sequential images.  
 
This work aims to infer individual droplets' trajectories by analyzing videos of microfluidic experiments using the latest computer vision tools. We combined the latest You Only Look Once 
(YOLOv5) algorithm for droplet recognition and the DeepSORT algorithm for 
droplet tracking in a single tool to achieve this goal. The object detector and 
tracking algorithms were tested previously  on noise-free image data generated by 
Lattice Boltzmann simulations~\cite{diotallevi2009} of dense emulsions~\cite{durve_epjp,durve_ptrsa}. 
Here, we extend the approach to the real images from microfluidic experiments significantly different from noise-free simulation images. The machine learning pipeline is tested on a microfluidic system producing clusters of densely packed, confluent droplets produced in a microfluidic setup reported in Bogdan et al. 2022~\cite{bogdan2022}. The system is important, among other reasons, because of intriguing similarities to naturally occurring clusters of cells. It also features previously unobserved dynamic modes and holds promise for future applications in tissue engineering and droplet libraries. 

In order to make a deep learning-based object detection and tracking pipeline, a generic object network training needs a large training dataset containing 
images of the desired objects. Training dataset compilation is a labor-intensive bottleneck involving manual image annotation of the desired objects. Here, we explore strategies of mixing synthetically prepared images and manually annotated real images to train the YOLO object detector networks. We find an optimal mixing proportion of synthetic and real data to gain the best performance and twofold reduction in manual image annotation efforts. The 
strategies explored in this work are not explicitly limited to the YOLO 
algorithm and can be applied to train other object detector networks. 


The paper is organized as follows. In the next section, we describe the experimental setup. In sec.~\ref{sec:alorithms} and sec.~\ref{sec:yolo_training}, we give a brief overview of the YOLO and the DeepSORT algorithms and describe the training procedure for various networks. In sec.~\ref{sec:result} we report our results on experimental videos with a focus on droplet counting accuracy of various networks and training data mixing strategies.

\section{experimental setup}
\label{exp_setup}

\begin{figure*}
\includegraphics [width = 13 cm ] {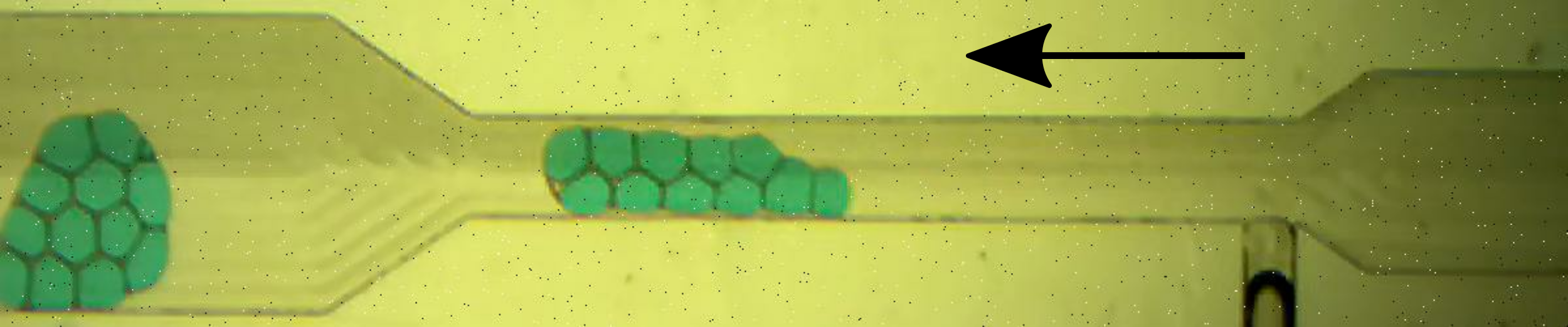}
\caption{ Clusters of droplets in a close-packed double emulsion~\cite{bogdan2022} translocate an orifice. The arrow indicates the flow direction. (Multimedia view) \label{label_fig1}}
\end{figure*}

The algorithm aims to successfully track fluid droplets enclosed in an external fluid under conditions demanding for traditional particle tracking methods. A specific group of systems, presenting a considerable challenge to automated droplet tracking, features flows of groups of very densely packed droplets, separated by a thin, lubricating layer of a mixture of oils with surfactants preventing the merging of droplets. In an 
experimental system of interest, these droplets are formed at a  T-junction of channels of rectangular cross-sections. Such generated monodisperse emulsion is pushed into a wider channel and broken up by the flow of an external phase into clusters of droplets in a flow-focusing junction. See Ref.~\cite{bogdan2022} for experimental details. This microfluidic system has been recently studied in the context of new dynamical modes in confined soft granular flows, such as stochastic jetting and dripping~\cite{bogdan2022}. An example of a cluster formed in such a system is shown in Fig.~\ref{label_fig1}. The fact that droplets fill up the space makes segmentation into separate droplets difficult. The droplets have various shapes, depending on forces acting from their neighbors. In addition, various coloring contrasts between the droplets, their interfaces, and their surrounding are used. The quality of the imagery may also differ between various experiments and researchers. 

Consequently, each experiment is documented in a video containing a few hundred images. Several commercial platforms, such as Matlab, offer video data analysis. However, the ready-to-deploy commercial platforms often cannot be easily customized for a unique dataset and may fail to analyze specific video instances. This paper aims to introduce state-of-the-art open-source algorithms and efficient data preparation strategies that can further be adapted for training object tracking ML pipelines. Subsequently, this work's trained networks can be used as a starting point for further enhancing the DropTrack using custom datasets.

\section{Algorithms}
\label{sec:alorithms}

A successful object tracking requires achieving two sub-goals, i) identifying the objects to track and ii) tracking the detected object in sequential frames. Typically, these two tasks are handled by two different algorithms, individually specialized in object recognition and object tracking. In this work, we employ the You only look once (YOLO) algorithm for object recognition and the DeepSORT algorithm for object tracking.

\subsection{You Only Look Once (YOLO)}

You Only Look Once (YOLO) is a single-stage object detection algorithm. A network trained with the YOLO algorithm identifies objects in a given image. The network output consists of bounding boxes around the detected objects, the class/category of each detected object, and the individual detection's confidence scores. The latest, the fifth version of YOLO (YOLOv5), is the fastest and most accurate object detector on two commonly used, general-purpose object detection 
datasets called Pascal VOC (visual object classes)~\cite{pascal} and Microsoft COCO object detection datasets~\cite{coco}. Many instances report inference speed of the YOLOv5 networks at or above 60 
frames per second (FPS)~\cite{zhou2021, luiz2021} for general object detection with various hardware configurations, and thus the YOLO object detector networks offer real-time image analysis. Due to their superior speed and accuracy, the YOLO networks are deployed for environmental  monitoring~\cite{forest_fire}, quality control 
processing~\cite{kiwifruit}, and checking protocol complience~\cite{covid_protocol} to name but a few applications.

The superior image analysis speed of the YOLO algorithm is achieved by its smart operating procedure~\cite{redmon, redmon1}. An input image is divided into $S \times S$ grids. Each cell is responsible for detecting object centroid present in the cell. Each grid cell then predicts $B$ bounding boxes with their confidence score for each detected object and $C$ conditional class probabilities for the given object belonging to a specific class. The predicted bounding boxes and class probabilities are combined to produce the final output as a single bounding box around the detected object and the class of that object. This final output is then passed to the object tracking DeepSORT algorithm.

\subsection{DeepSORT}

The DeepSORT algorithm is employed for tracking all detected objects between two successive frames~\cite{wojke}. By analyzing sequential frames, the DeepSORT can construct trajectories of all the detected objects. The DeepSORT algorithm works on two levels. At the first level, a classical Simple Online Real-Time tracking module~\cite{Bewley2016_sort} uses the Hungarian 
algorithm~\cite{hungarian} to distinguish detected objects in two 
consecutive frames and assigns individual objects their unique identity. Kalman filtering~\cite{kalman} is then used to predict the future position of the objects based on their current positions. At the second level, the deep network learns object descriptor features to minimize the identity switches as the object moves in subsequent frames. The YOLO and the DeepSORT algorithms together accomplish droplet recognition and tracking. 

In the next section, we outline the training data acquisition process employed in this work and later describe the training process for the YOLO network for droplet detection.

\section{Training the YOLO network}
\label{sec:yolo_training}

Object detector models are often employed to analyze real-world images. These images are highly complex, subject to various light conditions, camera noises, occlusions, and other environmental conditions. An object detector model needs to be trained with several images captured under a perceived broad range of environmental conditions for object recognition.

A typical object detector network training involves four key steps, selecting a network architecture, compiling training data, training a network, and testing the trained network's performance. Below, we describe these steps one by one in the context of the YOLOv5 network for droplet recognition.\\

\subsection{The YOLO network architectures}

\begin{figure*}
\includegraphics [width = 15 cm ] {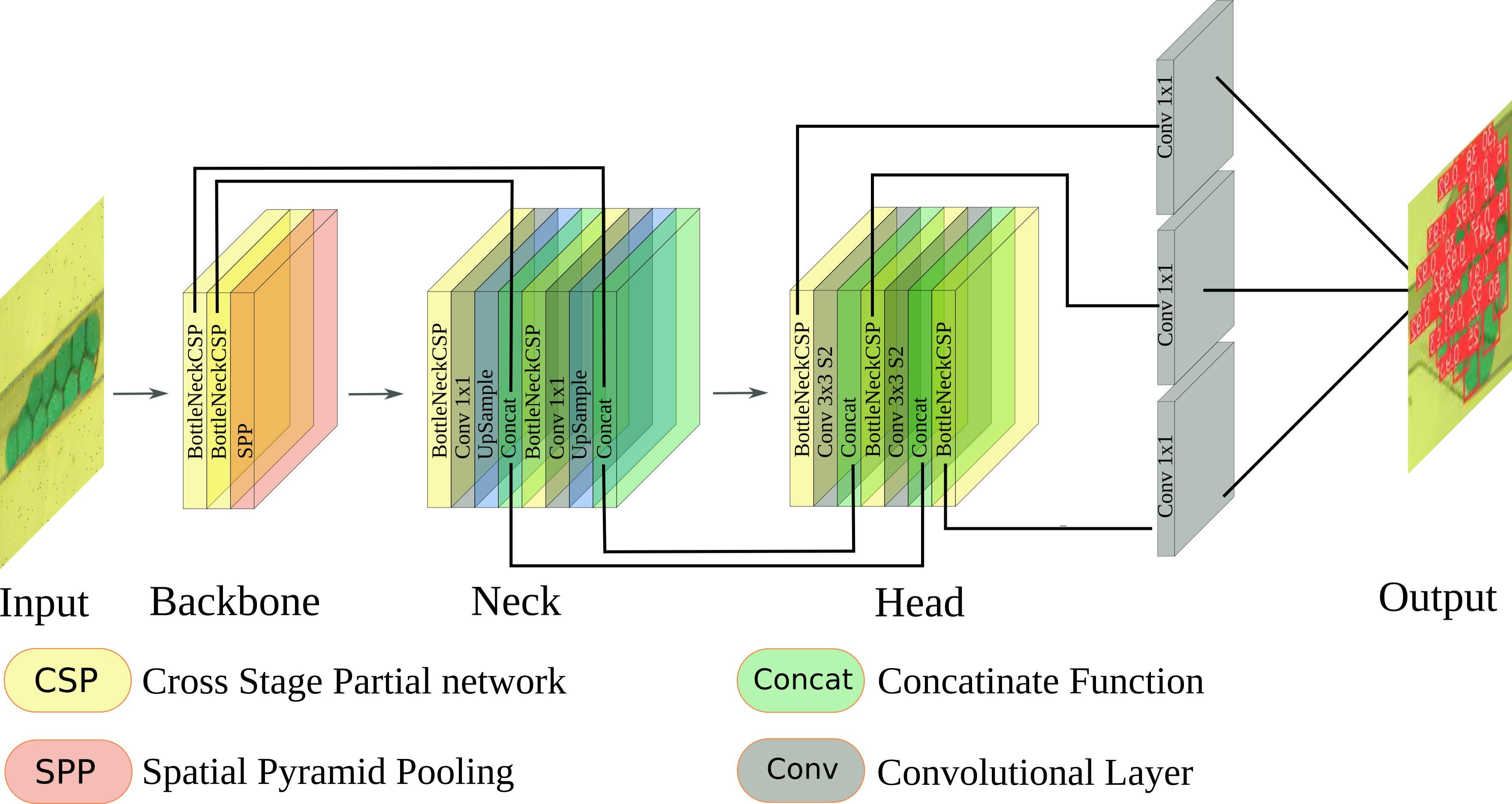}
\caption{ Sketch of the YOLOv5 network. The backbone extracts the feature from an input image. The neck does the feature propagation. The head network produces the output as bounding boxes around the detected droplets with confidence scores. The boxes in the figure represent deep layers of the YOLOv5 network. \label{label_fig2}}
\end{figure*}

The YOLOv5 network architecture is sketched in Fig.~\ref{label_fig2}. It consists of a cross-stage part network (CSPNet) as its backbone, whose architecture is responsible for the feature extraction from the input image. The neck part of the YOLOv5 network is made up of a path aggregation network (PANet), which improves localization signals in lower layers to enhance object localization accuracy. The last head part of the YOLOv5 network generates feature maps to locate objects across various scales. 

Table~\ref{YOLO_table} tabulates the details of several YOLO network architectures. The different networks differ in the number of tunable parameters. The smaller networks are faster but less accurate than the larger ones. This work trains eight YOLOv5 networks to identify droplets in microfluidic experimental videos and compare their performances with a measure for droplet counting error.

 \begin{table}
 \caption{\label{YOLO_table} Various YOLOv5 networks and their size in number of tunable parameters expressed in millions (M). }
 \begin{center}
\begin{tabular}{||c | c | c ||} 
 \hline
 Network name & Image size (pixels) &  \# parameters (M) 
\\ [0.5ex] 
 \hline\hline
  YOLOv5s & 640x640 & 7.2  \\
 \hline
 YOLOv5m & 640x640 & 21.2  \\
 \hline
 YOLOv5l & 640x640 & 46.5  \\
 \hline
 YOLOv5x & 640x640 & 86.7  \\  
 \hline
 YOLOv5s6 & 1280x1280 & 12.6 \\
 \hline
 YOLOv5m6 & 1280x1280 & 35.7 \\
 \hline
 YOLOv5l6 & 1280x1280 & 76.8 \\
 \hline
 YOLOv5x6 & 1280x1280 & 140.7 \\ [1ex] 
 \hline
\end{tabular}
\end{center}\end{table}

\subsection{Training data compilation}
The training dataset is a key component for training a YOLO network to detect objects of interest. Training data acquisition is a significant bottleneck in the quick development of object detector networks due to the laborious data gathering and annotation process. The training dataset is typically prepared by collecting several images captured under various conditions and then annotating these images. The image annotation process involves identifying and noting down the location of all the objects of interest in an associated label file. In general, image annotation is a highly labor-intensive process where human operators manually mark the objects in the images. Recently, it has been observed that the object detector networks could be trained with hybrid datasets containing real and fake images, thereby alleviating the labor-intensive image annotation work to some extent~\cite{saleh2018, wood2021}. In this work, we explore strategies of making a hybrid training dataset by mixing images taken from real microfluidics experiments and synthetically prepared ones using simple computer graphics tools. 

\begin{figure*}
\includegraphics [width = 13 cm ] {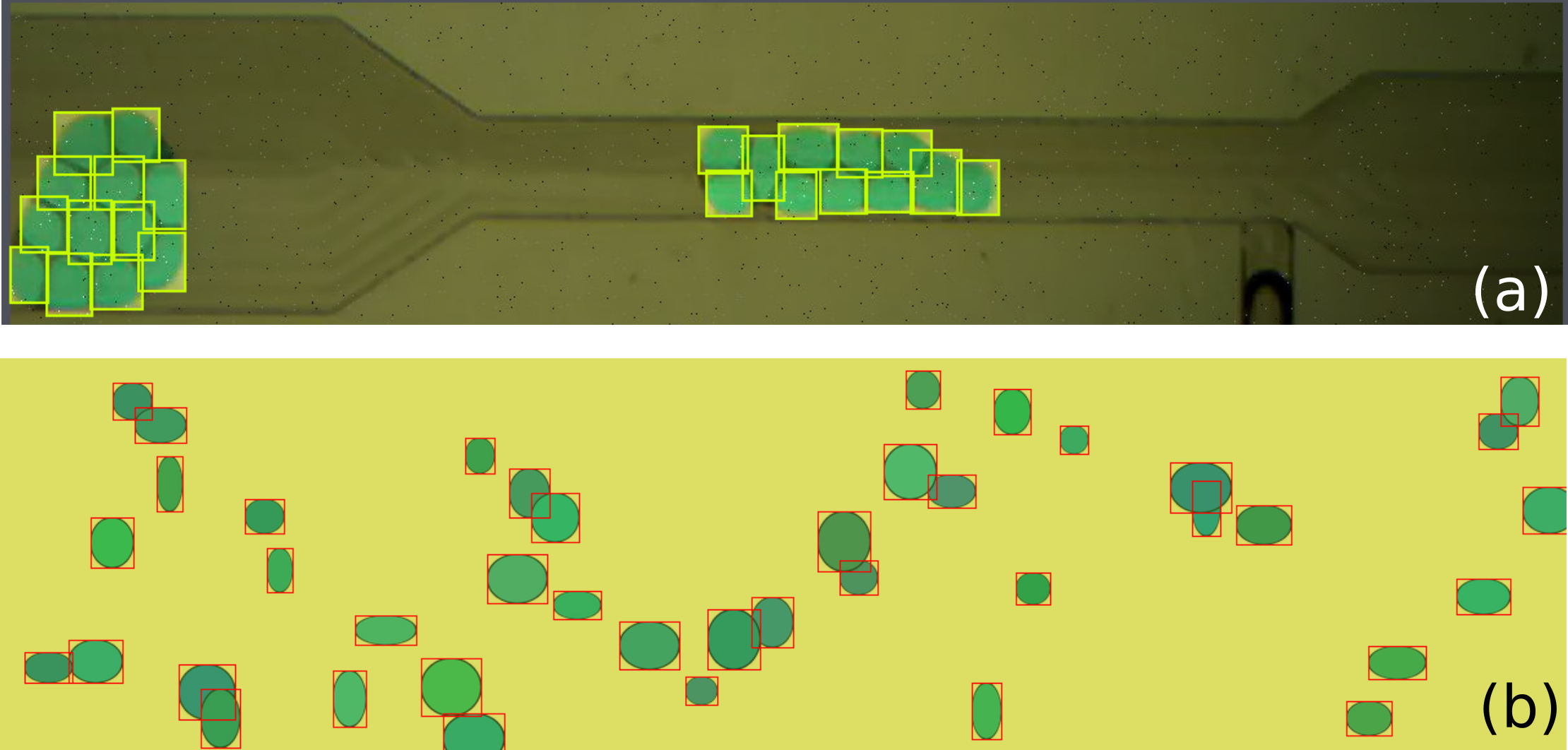}
\caption{ The training data set is composed of (a) manually labeled images from real experiments and (b) synthetically prepared images. Several 
training datasets were prepared with varying percentages of manual and 
synthetic images. The total number of images in these training datasets is always fixed at 800. 
\label{label_fig3}}
\end{figure*}

Hybrid training data used in this work to train the YOLO networks consists of two types of images. The first ones are the actual images taken from a single microfluidic experiment. The individual droplets in these images are manually annotated with the help of a free annotation tool Roboflow~\cite{roboflow}. A human operator draws the illuminated boxes around the droplets (see Fig.~\ref{label_fig3}(a)). For compiling a training dataset, in total, 800 frames were extracted from a video of a single experiment as described in Sec.~\ref{exp_setup}. The video that served as a training data source is included in Fig.~\ref{label_fig1} (Multimedia view). For simplicity, we call these images manually labeled images. The second type of images are prepared using computer graphics tools. In each of these images, a few solid ellipses mimicking the droplets are placed randomly on a uniform color background (see Fig.~\ref{label_fig3}(b)). The colors of solid circles and the background were varied randomly for each image to cover the wide color space observed in microfluidic droplet images. We call these images synthetic images. The hybrid training data consists of a mixture of these two types of images in various proportions, keeping the number of images in the training dataset constant at 800 images. In total, eleven datasets were prepared starting from 0\% synthetic data, i.e., all the images are real images, to 100\% synthetic images increasing in the steps of 10\% along the way.

All the eight networks mentioned in table~\ref{YOLO_table} were trained in separate training events, each involving a single network and a single training dataset out of the eleven training datasets. Thus, eighty-eight YOLO networks were trained to recognize droplets in the input images using the hybrid datasets described above. The details of the training procedure are described below. 

\subsection{Training results}

\begin{figure*}
\includegraphics [width = 13 cm ] {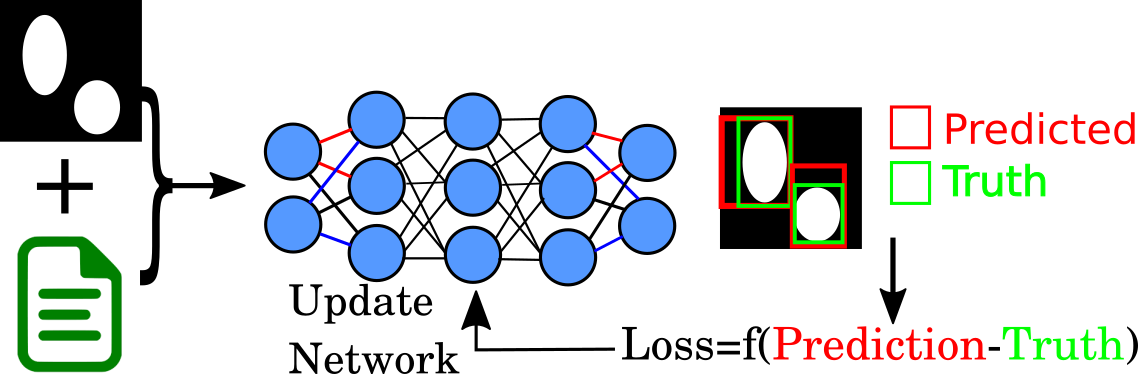}
\caption{ The YOLO network training scheme. Images and labels are fed to the network in batches, and the difference between the predictions and the ground truth is utilized in an iterative scheme to update network parameters for best performance.  
\label{label_fig4}}
\end{figure*}

The YOLO network is trained to identify objects of interest in an episodic setting. A subset of training data, called a batch, is given as input to the network in each episode. The data batch contains images and 
associated label files (see Fig.~\ref{label_fig4}). The network then outputs the predicted locations of the objects within these input images. The label files serve as a ground truth. A loss value for a batch of images is calculated using the difference between the ground truth, i.e., correct location and dimensions of the objects, and the network predictions for the same. The loss value is then used to update the network's tunable parameters to minimize the loss value for subsequent iterations. The YOLOv5 loss value is computed using bounding box regression loss, classification loss, and objectness loss~\cite{yolo_loss}. The bounding box regression loss accounts for wrong anchor box detection. It is a mean squared error on the predicted bounding box ($x, y, h, w$) and the given ground truth ($x', y', h', w'$). The classification loss is a cross entropy loss calculated for object classification, and the objectness loss is a mean squared error calculated for objectness confidence score~\cite{yolo_loss}.

The loss function values are used to tune the parameters of the model. However, they do not measure the model's actual performance in 
deployment conditions. The network's performance is measured during the training using a separately reserved validation dataset. After a predefined number of episodes, a validation dataset is given as input to the network, and its accuracy is measured as the mean average precision (mAP)~\cite{map_1} calculated from the precision and recall curves. The precision and recall values are computed with true-positive/negative and false-positive/negative detections. The true-positive/negative, false-positive/negative detections are determined by the intersection over union (IoU) of predicted bounding boxes and ground truth provided in the label file~\cite{Rezatofighi_2019_CVPR}. Thus, errors in manual image annotations affect the network training quality. At the end of the training, the optimized network, i.e., the values of tunable parameters, is stored in a separate file (called weights) that can be used for inference at later times.

The PyTorch implementation of the YOLO network training is open source and available online. We adopted an open-source code from this git repository~\cite{yolov5_git} for training the YOLOv5 networks. All the eight YOLO networks mentioned in table~\ref{YOLO_table} were trained separately with eleven hybrid datasets as described above. The hyperparameters used for the training are mentioned here~\cite{hyperparameters}. 
The training data contained 800 total images, with another 80 images reserved as a validation dataset. It is worth noting that the validation dataset contained only manually labeled images from real experiments to measure the network performance under the real deployment conditions.  

\begin{figure*}
\includegraphics[width = 14 cm, keepaspectratio]{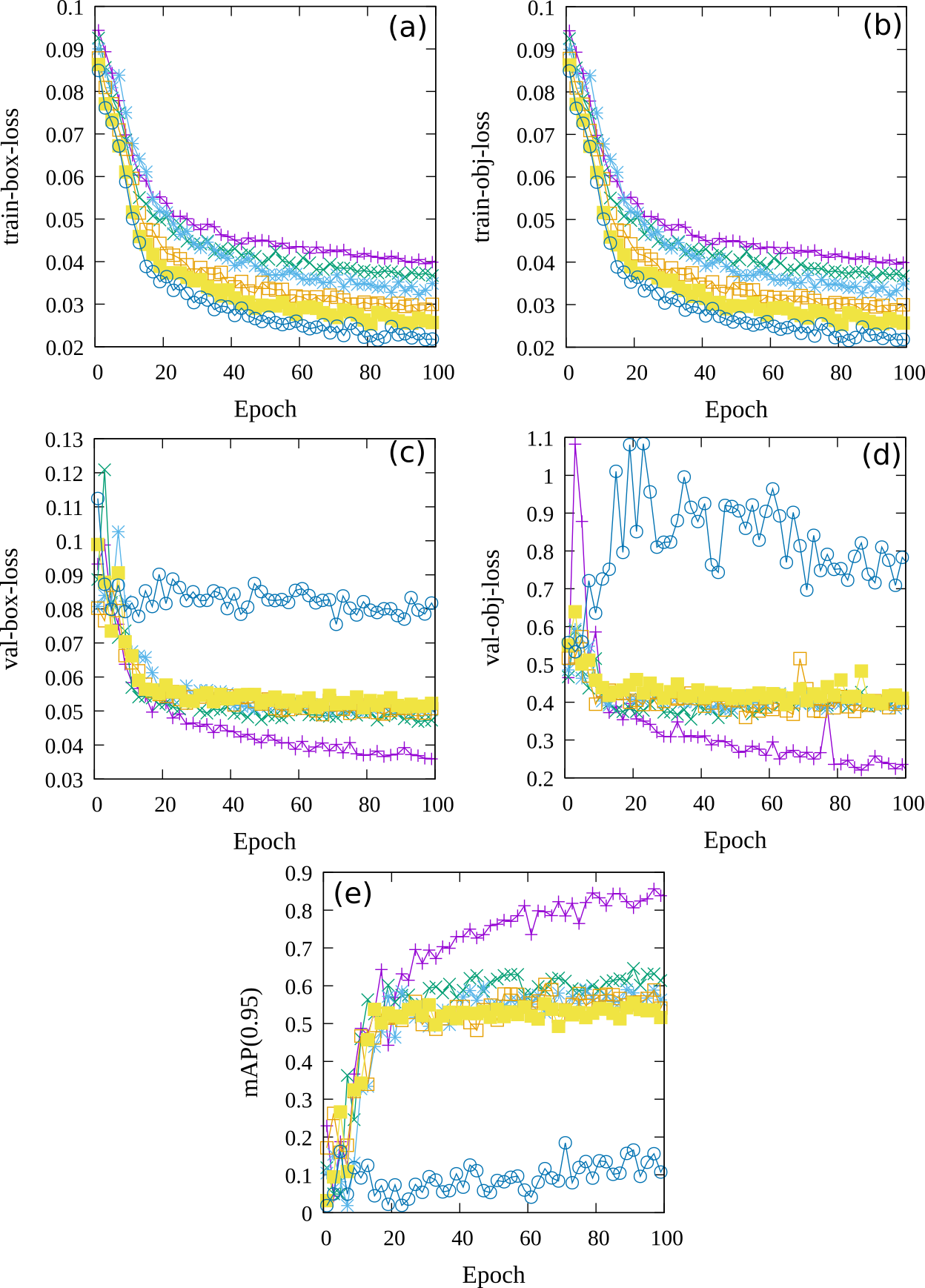}
\caption{ Loss associated with bounding boxes: (a) during the training, (c) 
during validation. Loss associated with object detection: (b) during the training, (d) during validation. (e) Mean average precision mAP as the training progress for various YOLO networks. For all sub-figures, the color codes are -- purple plus: 0\%, green cross: 20\%, blue asterisk: 40\%, orange squares: 60\%, yellow solid squares: 80\%, blue circles: 100 \% of synthetic data. \label{label_fig5}}
\end{figure*}

In Fig.~\ref{label_fig5} we show regression loss and objectness loss values during the training and validation for the YOLOv5l network for various training data compositions. The YOLOv5l network's training profile represents all other training events as they show similar trends. Since the network is being trained to detect only a single class, classification loss measurements are not shown. Various indicators measure training progress. Figure~\ref{label_fig5} (a) and (c) show training box loss as the training progress for training and validation data, respectively. The training box loss is associated with an error in the bounding box prediction of the YOLO network. Figure~\ref{label_fig5} (b) and (d) show loss associated with object recognition during the training progress. As the training progresses, bounding box regression values and object loss values decrease and saturate to a value for all training datasets. However, three distinct patterns emerged for the loss values computed on the validation datasets. The loss values for purely synthetic images (blue circles in Fig.~\ref{label_fig5} (c) and (d)) remain comparatively high throughout the training, whereas the loss values for purely manually labeled images (purple plus in Fig.~\ref{label_fig5} (c) and (d)) decrease from high starting values and later saturate to minimum values. There is no significant difference in their loss saturation values for all other in-between datasets. A similar trend is observed for the network evaluation with mAP values. The network trained with a dataset containing only the manually labeled images achieves the highest mAP, which indicates the best performance. In contrast, the network trained with a dataset containing only the synthetic images has the lowest mAP score, indicating the worst performance on the validation dataset. All other networks attain almost identical mAP scores indicating that these models have comparable performance. The trained YOLOv5 networks are then combined with the DeepSORT and deployed to analyze videos from actual experiments.

\section{Results}
\label{sec:result}
\begin{figure*}
\includegraphics [width = \textwidth]{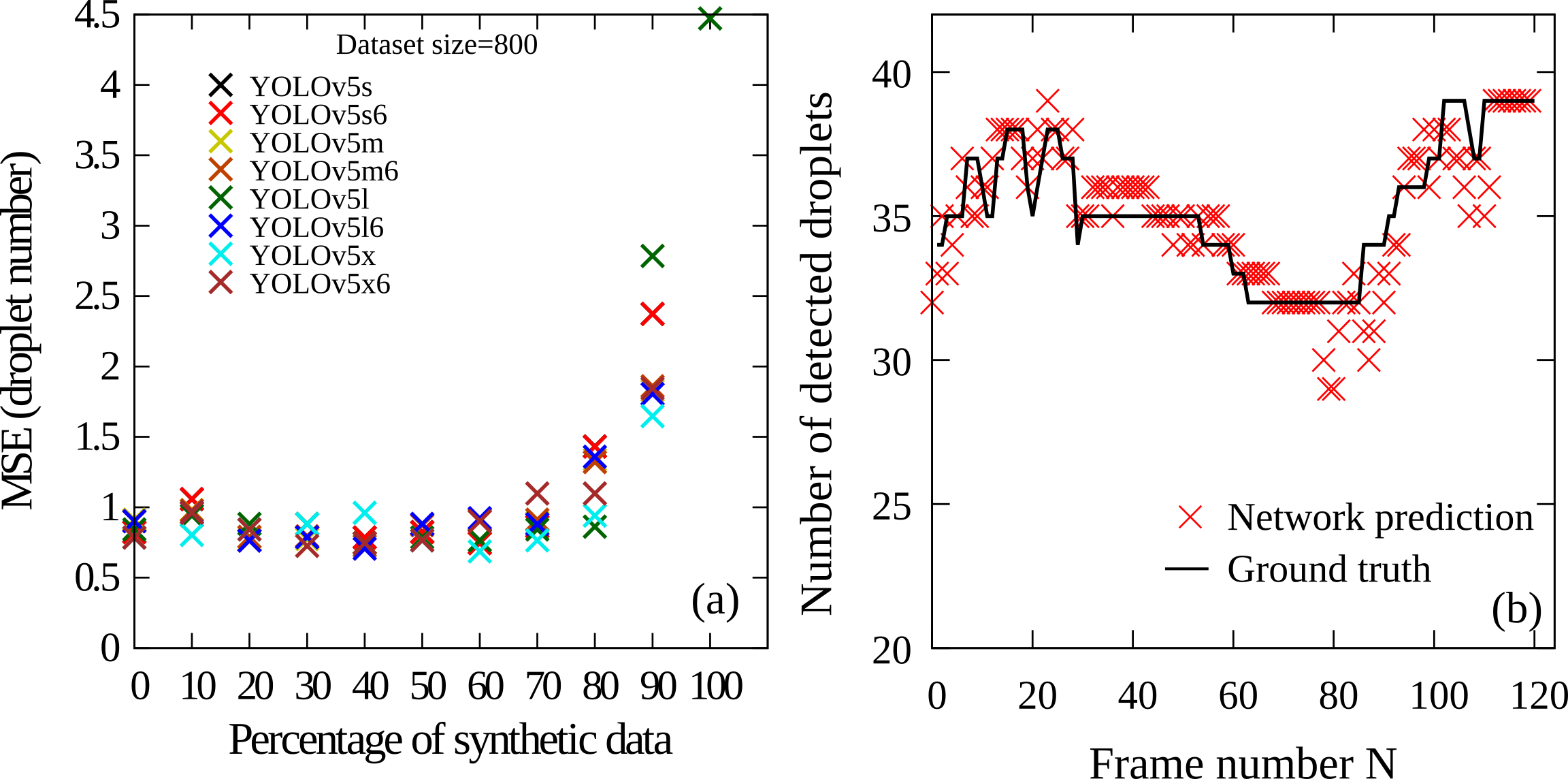}
\caption{ Measuring accuracy of the trained YOLO networks. (a) mean square error (MSE) between predicted droplet number and ground truth for respective YOLO networks trained with different proportions of manual and synthetic data. (b) Number of droplets detected by the YOLOv5l network compared with the manually counted number when 
trained with a training dataset consisting of 60\% synthetic data and 40\% manually labeled data. 
\label{label_fig6}}
\end{figure*}

Below, we present inference results on a video from the microfluidic experiment. The analyzed video was not a part of the training dataset and served as a test case. The test case video contained more droplets and clusters than the one used for training the YOLO networks, thus presenting more complexity for the trained YOLO networks. Here, we analyze the network outputs for errors in droplet number counting, which is one of the basic desired microfluidic experimental video data 
analyses. The test video contained 120 frames. All the droplets in it were manually counted to measure the YOLO networks' accuracy. Figure~\ref{label_fig6}(b) shows the manually counted droplet numbers and the YOLOv5l network's prediction for every frame in the test video. The trained YOLOv5 networks' accuracy was defined as the mean squared error (MSE) between the detected droplets in each frame and manually counted droplet numbers serving as a ground truth. The MSE is computed as $MSE = \frac{1}{F_n} \sum_{i=1}^{F_n}\sqrt{(M_i - P_i)^2} $,  where $M_i$ is the number of manually counted droplets in the $i^{\text{th}}$ frame, $P_i$ is the number of droplets located by the YOLOv5 network in the $i^{\text{th}}$ frame, and $F_n$ are the total number of frames in the test video. Figure~\ref{label_fig6}(a) shows MSE for the YOLO networks trained with a varied proportion of manual and synthetic images in the training dataset. It is clear from Fig.~\ref{label_fig6}(a) that there is no significant difference in the accuracy between networks trained with 0\% of synthetic data and networks trained with up to 60\% of synthetic data, thereby opening a prospect of significant savings on manual labeling efforts. It is worthwhile to note that the 60\% synthetic data and 40\% manually labeled data combination is not claimed to be a universal feature for all object detector models. Nevertheless, this result shows that a combination of real and synthetic data in an optimal proportion can significantly reduce manual efforts without any noticeable performance loss.    


\begin{figure*}
\includegraphics [width = 13 cm ] {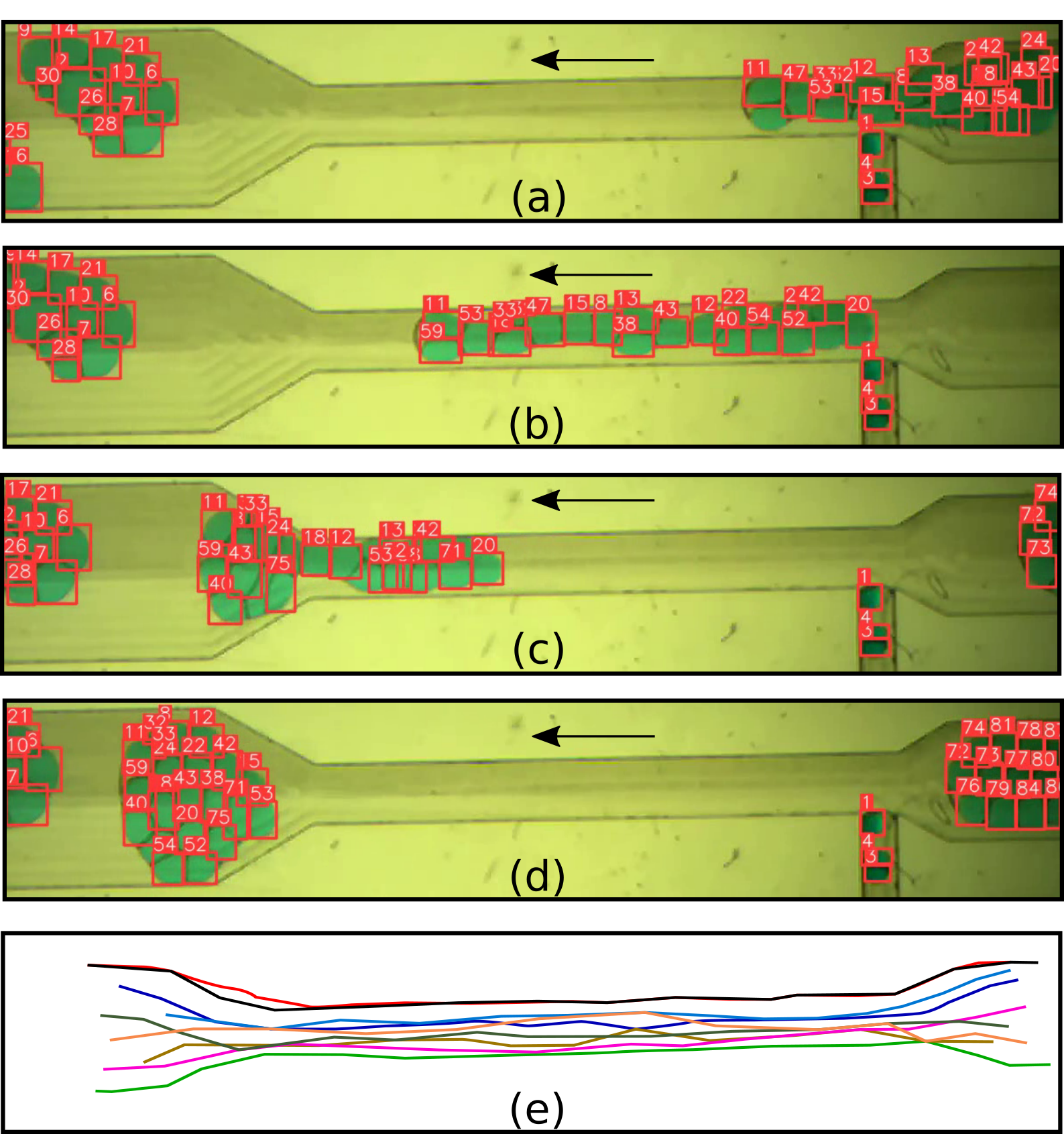}
\caption{Snapshot of the analyzed video at various frame number N (a) N=30, (b) N=60, (c) N=90, (d) N=120. The YOLO network predicts the red bounding boxes around each droplet, and the DeepSORT assigns a unique number to each droplet. The arrows indicate the flow direction. The video is analyzed with a model trained with 60\% synthetic data and 40\% manual data. (e) Trajectories of successfully tracked individual droplets are shown with unique colors. Other trajectories (not shown here) were generated in parts due to ID switching, which can be glued with heuristic algorithms. (Multimedia view)
\label{label_fig7}}
\end{figure*}

Finally, combining the optimal YOLO network with DeepSORT, trajectories of the individual droplets are traced in a video. We adapted open-source YOLOv5 and DeepSORT code~\cite{yolov5_deepsort_git} to run the inference. The snapshots of droplet identification and tracking are shown in Fig.~\ref{label_fig7}(a)-(d). The YOLO algorithm predicts the red bounding boxes, and the DeepSORT algorithm achieves the tracking of these droplets by assigning unique IDs to the droplets. The full tracking video is included in Fig.~\ref{label_fig7} (Multimedia view). Incidentally, 19 droplets pass through the narrow channel. DropTrack constructs nine full trajectories, 
meaning these full trajectories trace the droplets as they enter the channel until they leave. The trajectories are constructed in parts for other droplets due to erroneous droplet ID switching by the DeepSORT algorithm as the droplets move and deform. In Fig.~\ref{label_fig7}(e), we show only the full trajectories of the center of mass of the successfully tracked droplets that pass through the narrow constriction. The center of mass of a droplet is approximated as the center of its bounding rectangle. Further research is needed to achieve the robust tracking of deformable objects with the DeepSORT before DropTrack can be deployed as a real-time and perfect tracking tool for objects such as droplets, cells, etc.     

Image analysis speed is crucial for real-time applications, along with detection and tracking accuracy. Table~\ref{inference_table} shows the time taken to analyze an input image by the YOLOv5 and the DeepSORT algorithms using an Intel Xeon CPU and NVIDIA V-100 GPU machines. On both the CPU and GPU implementations, the larger YOLOv5 networks take more time to analyze a given input than smaller YOLOv5 networks. On the other hand, the DeepSORT algorithm takes comparable time as the network size is unchanged. Note that the GPU implementation is about 100x faster than the CPU implementation. The GPU image analysis rate (measured in frames per second FPS) is good enough for real-time droplet tracking. 

\begin{table}
 \caption{\label{inference_table} Individual YOLO and DeepSORT inference time for various networks compared on a single core of Intel Xeon Platinum 8176 CPU @ 2.80 GHz and high-end NVIDIA V-100 GPU. Inference time is measured in seconds and the inference speed is calculated as frames per second (FPS) rate for DropTrack. The GPU implementation is two orders of magnitude faster than the CPU implementation.  }
 \begin{center}
\begin{tabular}{||c | c | c | c | c| c | c ||} 
 \hline
 Network name  & \makecell{ YOLO \\ (Xeon CPU)} & 
 \makecell{DeepSORT \\(Xeon CPU)} & \makecell{Total FPS \\ (CPU)} &  \makecell{YOLO \\(V100-GPU)} &  \makecell{DeepSORT \\ (V100-GPU)} & \makecell{Total FPS \\ (GPU)} \\ [0.5ex] 
 \hline\hline
  YOLOv5s + DeepSORT &0.104 s &7.882 s & 0.13 & 0.007 s & 0.026 s & 30.30 \\
 \hline
 YOLOv5m + DeepSORT &0.227 s  &8.574 s & 0.11 &0.009 s & 0.027 s & 27.78\\
 \hline
 YOLOv5l + DeepSORT & 0.439 s &8.641 s &0.11 & 0.010 s & 0.027 s & 27.03\\
 \hline
 YOLOv5x + DeepSORT & 0.797 s &8.412 s &0.11 & 0.013 s & 0.028 s & 24.39\\  
 \hline
 YOLOv5s6 + DeepSORT &0.309 s  &8.676 s & 0.11  & 0.009 s & 0.029 s & 26.32\\
 \hline
 YOLOv5m6 + DeepSORT &0.719 s &8.159 s &0.11 & 0.012 s & 0.030 s & 23.81 \\
 \hline
 YOLOv5l6 + DeepSORT & 1.560 s &8.441 s &0.11 & 0.014 s & 0.029 s & 23.26\\
 \hline
 YOLOv5x6 + DeepSORT & 2.660 s &8.554 s &0.09 & 0.017 s & 0.030 s & 21.28 \\ [1ex] 
 \hline
\end{tabular}
\end{center}\end{table}

\section{Conclusion}
This work combines the YOLOv5 object detector and the DeepSORT object tracker for the automatic detection and tracking of droplets in microfluidic experiments. Several YOLOv5 networks were trained with a mix of synthetic and real images. The synthetic images were generated with computer graphics tools, while the real images were taken from actual microfluidic experiments and were manually annotated. For our test case, the network trained with by combining 60\% synthetic data and 40\% real data show similar droplet detection performance (in terms of counting the droplets) compared with the network trained only with real data, meanwhile significantly reducing manual image annotation work by a factor two. These
results highlight the effectiveness of hybrid datasets in training an object 
detector network and provide a valuable strategy to save significant resources while developing similar applications to analyze challenging bio/microfluidic experimental data. 
   
Furthermore, this work presents a case study for object detection and tracking in a challenging scenario due to compact and deformable droplet configurations. We report inference speed and a measure for training quality, mean average precision for several YOLOv5 networks. A trained network with 60\% training data with relatively low mAP performs similarly to the network with a high mAP value (100\% real data) in droplet counting. This observation highlights the need to check for various network training quality measures, which could be tailored to a specific application and hints toward diversifying reliance on a standard measure (mean average precision) used to measure object detector accuracy.

Due to identity switches by the DeepSORT, the present tool could successfully track about half of the droplets that undergo significant deformations and accelerated motions. The trajectories of the remaining droplets were generated as partial segments that could be glued together with heuristic algorithms. Further exciting research direction is to consolidate state-of-the-art algorithms to ensure the best object detection, minimum identity switches, and continuous trajectories of highly deformable objects such as droplets, biological organisms, and cells for complete and real-time tracking.

\section{Acknowledgement}
The authors acknowledge funding from the European Research Council under the 
European Union's Horizon 2020 Framework Programme (No. FP/2014-2020) ERC Grant 
Agreement No.739964 (COPMAT).  J. G. acknowledges support from Foundation for 
Polish Science within First Team program under Grant 
No. POIR.04.04.00-00-26C7/16-00. M. B. acknowledges the PMW program of the 
Minister of Science and Higher Education in the years 2020-2024 No. 
5005/H2020-MSCA-COFUND/2019/2. We gratefully acknowledge the HPC infrastructure 
and the Support Team at Fondazione Istituto Italiano di Tecnologia.

\section{Data availability statement}
The data that supports the findings of this study are available within the 
article and its supplementary material.

\section{Author Declarations}
The authors have no conflicts to disclose. The following article has been 
submitted to Physics of Fluids.

\bibliography{Ref}

\end{document}